\documentclass{article}
\usepackage{spconf,amsmath,amsfonts,graphicx,cite}

\def\*{\textsuperscript{$\!\ast$}}

\DeclareMathOperator{\tr}{Tr}

\renewcommand{\c}{{\bf c}}

\newcommand{\vvec}{{\bf v}}

\newcommand{\x}{{\bf x}}

\newcommand{\z}{{\bf z}}

\newcommand{\A}{{\bf A}}
\newcommand{\B}{{\bf B}}
\newcommand{\C}{{\bf C}}

\newcommand{\I}{{\bf I}}

\newcommand{\M}{{\bf M}}

\newcommand{\R}{\mathbb{R}}

\newcommand{\W}{{\bf W}}
\newcommand{\X}{{\bf X}}

\newcommand{\Z}{{\bf Z}}

\title{An Online Algorithm for Contrastive Principal Component Analysis}
\name{Siavash Golkar\thanks{$^\ast$SG and DL contributed equally.}$^{\ast,\dag}$, David Lipshutz$^{\ast,\dag}$, Tiberiu Tesileanu$^\dag$, Dmitri B.\ Chklovskii$^{\dag,\ddag}$}
\address{$^\dag$Center for Computational Neuroscience, Flatiron Institute \\ $^\ddag$Neuroscience Institute, NYU Langone Medical School}
\begin{document}
%
\maketitle
\begin{abstract}
Finding informative low-dimensional representations that can be computed efficiently in large datasets is an important problem in data analysis. Recently, contrastive Principal Component Analysis (cPCA) was proposed as a more informative generalization of PCA that takes advantage of contrastive learning. However, the performance of cPCA is sensitive to hyper-parameter choice and there is currently no online algorithm for implementing cPCA. Here, we introduce a modified cPCA method, which we denote cPCA\*, that is more interpretable and less sensitive to the choice of hyper-parameter. We derive an online algorithm for cPCA\* and show that it maps onto a neural network with local learning rules, so it can potentially be implemented in energy efficient neuromorphic hardware. We evaluate the performance of our online algorithm on real datasets and highlight the differences and similarities with the original formulation.
\end{abstract}
\section{Introduction}
\label{sec:intro}




Principal Component Analysis (PCA) is a long-standing pillar of dimensionality reduction that is often a first step in data analysis. As a spectral method without hyper-parameters, PCA is robust and interpretable. Furthermore, there exist efficient algorithms for computing the principal subspace projections \cite{allen2017first}, including online algorithms that map onto networks with local learning rules and can potentially be implemented in energy efficient neuromorphic hardware \cite{pehlevan2015hebbian}.

However, PCA projects a dataset onto the directions of highest variance, without regard to the source of this variability. In particular, if there are directions with high variability due to noise, PCA projects onto those directions, potentially at the cost of more informative subspaces.
Recently, contrastive PCA (cPCA) was proposed as a generalization of PCA that can account for such scenarios~\cite{cpca,abid2018exploring}. 
By using a background dataset comprised of domain-relevant `negative samples' (e.g., a music studio recording sans music), cPCA can distinguish directions of interest from those that have high variance due to background noise. 
For example, in a drug trial, cPCA picks an informative subspace by discounting the directions of high variance in the control (or placebo) group. This method has proved useful for both analysis and visualization in domains where a background dataset is available~\cite{covid_or_flu, marchetti2020structured, supporting}. It has also been applied to the image and video compression domain~\cite{image/video}.

While cPCA has been shown to find more informative projections than PCA~\cite{cpca}, this improvement has come at the cost of interpretability and efficiency. 
In particular, cPCA is highly sensitive to a hyper-parameter that sets the relative contribution of the positive and negative samples. This renders the algorithm less robust and also less interpretable than PCA. Furthermore, online PCA algorithms are not readily adapted to perform cPCA, so cPCA is less amenable to large datasets or in scenarios with limited amount of compute. 


In this work, we introduce cPCA\*, a more robust variation of cPCA.
We show cPCA\* is less sensitive to hyper-parameter choice and, for specific generative examples, can be interpreted as finding the subspace that maximizes the signal-to-noise ratio. We derive an online algorithm for cPCA\* and map it onto a neural network with local learning rules which can be implemented on efficient neuromorphic chips. In summary,
\begin{itemize}
    \item We introduce cPCA\*, an interpretable contrastive PCA method, and derive an online algorithm that maps onto a neural network with local learning rules.
    \item We empirically show that cPCA\* is more robust to the choice of hyper-parameter than cPCA.
\end{itemize}

\section{\lowercase{c}PCA: original and our method}

Given a dataset consisting of positive and negative samples, we want to identify directions that have large variance in the positive samples but small variance in the negative samples.

To be precise, let $d\ge2$ and $(\x_1,\delta_1),\dots,(\x_T,\delta_T)\in\R^d\times\{0,1\}$ be a sequence of inputs. The input $\x_t$ is a feature vector, which can either be a \textit{positive sample} or a \textit{negative sample}. The scalar $\delta_t$ is an indicator variable such that $\delta_t=1$ for positive samples and $\delta_t=0$ for negative samples.
We define covariance matrices of the positive and negative samples, as follows:
\begin{align*}
    \C_{(+)}:=\langle\x_t\x_t^\top|\delta_t=1\rangle_t,&&\C_{(-)}:=\langle\x_t\x_t^\top|\delta_t=0\rangle_t.
\end{align*}
cPCA finds directions (i.e., unit vectors $\vvec\in\R^d$) that simultaneously maximize $\vvec^\top\C_{(+)}\vvec$ and minimize $\vvec^\top\C_{(-)}\vvec$.

\subsection{Contrastive PCA.}
Abid et al.\ \cite{cpca,abid2018exploring} proposed projecting the positive samples onto the top $k$-dimensional eigen-subspace of the matrix difference 
\begin{align}
    \A_\alpha:=(1-\alpha)\C_{(+)}-\alpha\C_{(-)},\qquad\alpha\in[0,1],
\end{align}
where $1\le k<d$ and $\alpha$ is the \textit{contrast parameter}.\footnote{Abid et al.\ \cite{abid2018exploring} find the top eigen-subspace of $\C_{(+)}-\alpha\C_{(-)}$ for $\alpha\ge0$. We reparametrize the contrast parameter for a more direct comparison with our method.}

To gain intuition, consider the special case where positive samples correspond to noisy signal and negative samples are pure noise; that is,
\begin{align}\label{eq:signal}
    \x_t=\delta_t\times\text{signal}+\text{noise}.
\end{align}
If $\delta_t=1$, then $\x_t$ is signal plus noise, whereas if $\delta_t=0$, then $\x_t$ is just noise. Under the assumption that the signal and noise are decorrelated, the covariance of the signal is equal to $2\A_{1/2}=\C_{(+)}-\C_{(-)}$, so the directions that maximize the variance of the signal are the top eigenvectors of $\A_{1/2}$.

The contrast parameter $\alpha$ represents the trade-off between maximizing the target variance and minimizing the background variance. 
When $\alpha=0$, cPCA 
reduces to PCA applied to the positive samples. 
As $\alpha\to1$, directions that reduce the variance of the negative samples become more optimal and the contrastive principal subspace is driven towards the minor subspace of the negative samples.
Therefore, each value of $\alpha$ yields a direction with a different trade-off between positive and negative samples variance. 

In practice, the performance of cPCA is sensitive to the contrast parameter $\alpha$ (see Fig.~\ref{fig:mnist_hyper}). While the generative model suggests that it should perform optimally when $\alpha=1/2$, in practice, cPCA often performs better for $\alpha>1/2$ (Fig.~\ref{fig:mnist_hyper}) and there is no theory for why this is true. Furthermore, since $\A_\alpha$ is not a covariance matrix for $\alpha\in(0,1)$, online PCA algorithms \cite{allen2017first,pehlevan2015hebbian} cannot be adapted to cPCA.

\subsection{Our method: Contrastive PCA\*.}
Our method, cPCA\*, is based on projecting the positive samples onto the top $k$-dimensional eigen-subspace of the generalized eigenvalue problem
\begin{align}\label{eq:GEV}
    \C_{(+)}\vvec=\lambda\B_\beta\vvec,
\end{align}
where
\begin{align}\label{eq:B}
    \B_\beta:=(1-\beta)\I_d+\beta\C_{(-)},\qquad\beta\in[0,1],
\end{align}
and the hyper-parameter $\beta$ is our \textit{contrast parameter}.

To motivate cPCA\*, consider the generative model in Eq.~\eqref{eq:signal}, under the assumption that the signal and noise are uncorrelated. Let $\C_\text{signal}=\C_{(+)}-\C_{(-)}$ and $\C_\text{noise}=\B_1=\C_{(-)}$. Then the direction (i.e., unit vector $\vvec$) that maximizes the signal-to-noise ratio is the unit vector $\vvec$ that maximizes the ratio
\begin{align*}
    \frac{\vvec^\top\C_\text{signal}\vvec}{\vvec^\top\C_\text{noise}\vvec}=\frac{\vvec^\top\C_{(+)}\vvec}{\vvec^\top\B_1\vvec}-1.
\end{align*}
The optimal $\vvec$ corresponds to the top eigenvector of the generalized eigenvalue problem \eqref{eq:GEV} with $\beta=1$. 

As with the cPCA contrast parameter $\alpha$, when $\beta=0$, the problem reduces to PCA applied to positive samples. As $\alpha\to1$ in cPCA, the projection subspace converges to the negative sample minor subspace, whereas as $\beta\to1$ in cPCA\*, the projection converges to maximize the ratio of the variance of the positive sample projections and the variance of the negative sample projections. For $\beta\in(0,1)$, our method is interpretable as maximizing the ratio of the variance of the positive sample projections and the variance of the negative sample projections when the negative samples have been conditioned with independent isotropic noise with variance $\tfrac{1-\beta}{\beta}$.

\section{Online \lowercase{c}PCA\* algorithm}

To derive an online algorithm, we concatenate the sequence of products $\delta_1\x_1,\dots,\delta_T\x_T\in\R^d$ into the data matrix:
\begin{align*}
    \X_{(+)}:=[\delta_1\x_1,\dots,\delta_T\x_T]\in\R^{d\times T}.
\end{align*}
Since $\delta_t\x_t$ is zero for negative samples and equal to $\x_t$ for positive samples, the covariance of $\X_{(+)}$ is a scalar multiple of the covariance of the positive samples $\C_{(+)}$.

\textbf{Similarity matching objective.}
Our starting point is the similarity matching objective:
\begin{align}\label{eq:SM}
    \min_{\Z\in\R^{k\times T}}\frac{1}{T^2}\|\Z^\top\Z-\X_{(+)}^\top\B_\beta^{-1}\X_{(+)}\|_F^2.
\end{align}
The similarity matching objective, which is a special case of a cost function from multidimensional scaling \cite{cox2008multidimensional}, minimizes the difference in similarity (measured using inner products) between the outputs $\Z$ and the positive samples $\X_{(+)}$ normalized by $\B_\beta$. The optimal solution $\Z$ of equation \eqref{eq:SM} is the projection of $\B_\beta^{-1/2}\X_{(+)}$ onto its $k$-dimensional principal subspace \cite{cox2008multidimensional}, which is equal to the projection of $\X_{(+)}$ onto the top $k$-dimensional eigen-subspace of the generalized eigenvalue problem \eqref{eq:GEV}.

\textbf{Legendre transforms.}
Directly optimizing Eq.~\eqref{eq:SM} does not result in an online algorithm. Rather, we follow the approach in \cite{lipshutz2020biologically,lipshutz2021biologically} for deriving an online algorithm. Expanding the square in Eq.~\eqref{eq:SM} and dropping terms that do not depend on $\Z$ yields the minimization problem
\begin{align*}
    \min_{\Z\in\R^{k\times T}}\frac{1}{T^2}\tr\left(\Z^\top\Z\Z^\top\Z-2\Z^\top\Z\X_{(+)}^\top\B_\beta^{-1}\X_{(+)}\right).
\end{align*}
We substitute in with the Legendre transforms:
\begin{align*}
    &\frac{1}{T^2}\tr\left(\Z^\top\Z\Z^\top\Z\right)=\max_{\M\in\mathbb{S}_{++}^k}\frac2T\tr\left(\Z^\top\M\Z\right)-\tr\left(\M^2\right),
\end{align*}
where $\mathbb{S}_{++}^k$ is the set of $k\times k$ positive definite matrices, and
\begin{align*}
    &\frac{1}{T^2}\tr\left(\Z^\top\Z\X_{(+)}^\top\B_\beta^{-1}\X_{(+)}\right)\\
    &\qquad=\max_{\W\in\R^{k\times d}}\frac2T\tr\left(\Z^\top\W\X_{(+)}\right)-\tr\left(\W\B_\beta\W^\top\right),
\end{align*}
which are respectively optimized at $\M^\ast=\frac1T\Z\Z^\top$ and $\W^\ast=\frac1T\Z\X_{(+)}^\top\B_\beta^{-1}$.
After substitution, we have
\begin{align*}
    \min_{\Z\in\R^{k\times T}}\min_{\W\in\R^{k\times d}}\max_{\M\in\mathbb{S}_{++}^k}L(\W,\M,\Z),
\end{align*}
where
\begin{align*}
    L(\W,\M,\Z)&:=\frac1T\tr(2\Z^\top\M\Z-4\Z\W\X_{(+)})\\
    &\qquad-\tr(\M^2+2\W\B_\beta\W^\top).
\end{align*}
Since the objective satisfies the saddle-point property with respect to $\M$ and $\Z$, we can interchange the order of optimization:
\begin{align}\label{eq:minmax}
    \min_{\W\in\R^{k\times d}}\max_{\M\in\mathbb{S}_{++}^k}\min_{\Z\in\R^{k\times T}}L(\W,\M,\Z),
\end{align}

\textbf{Offline algorithm.}
Before deriving an online algorithm, it is instructive to first solve Eq.~\eqref{eq:minmax} in the offline setting where we have access to the dataset $\X_{(+)}$ and the matrix $\B_\beta$. In this setting, we first minimize $L(\W,\M,\Z)$ with respect to $\Z$:
\begin{align*}
    \Z=\M^{-1}\W\X_{(+)}.
\end{align*}
We then take gradient descent-ascent steps:
\begin{align}\label{eq:deltaW}
    \W&\gets\W+2\eta\left(\frac1T\Z\X_{(+)}^\top-\W\B_\beta\right)\\ \label{eq:deltaM}
    \M&\gets\M+\frac\eta\tau\left(\frac1T\Z\Z^\top-\M\right).
\end{align}
Here $\tau>0$ denotes the ratio between the updates to $\W$ and $\M$ (which we set to 1 in our experiments) and $\eta\in(0,\tau)$ is the step size.

\begin{figure}
    \centering
    \includegraphics[width=.23\textwidth]{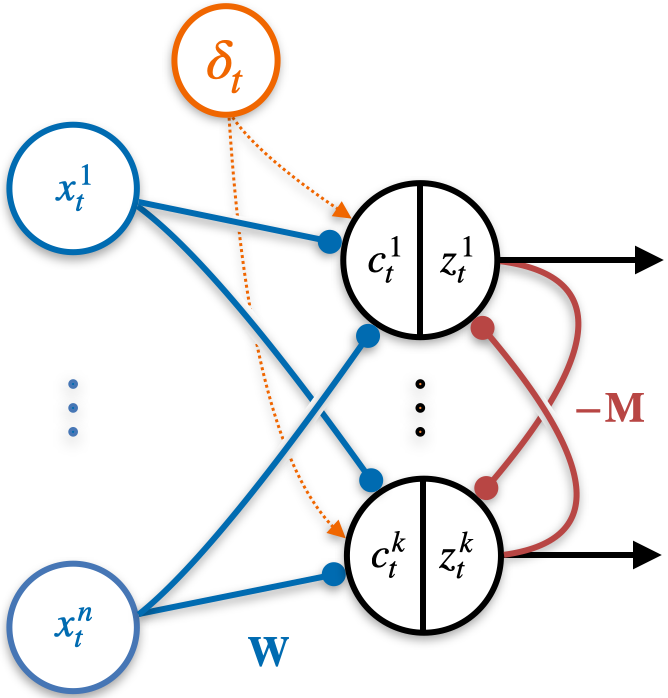}
    \caption{Neural network implementation of our online cPCA\* algorithm. At each $t$, the inputs $(x_t^1,\dots,x_t^d)$ are projected onto the feedforward weights $\W$ to obtain $(c_t^1,\dots,c_t^k)$. Lateral weights $-\M$ connect the output neurons. The neural dynamics in Eq.~\eqref{eq:neural} are run to equilibrium and the output of the network is $(z_t^1,\dots,z_t^k)$. The synaptic weights $\W$ and $-\M$ are then updated according to Eqs.~\eqref{eq:W} and \eqref{eq:M}.}
    \label{fig:nn}
\end{figure}

\textbf{Online algorithm.}
To solve the minimax objective \eqref{eq:minmax} in the online setting, we take stochastic gradient descent-ascent steps. At each time step $t$, we first minimize over the output $\z_t$ by running the continuous dynamics to equilibrium:
\begin{align}\label{eq:neural}
    \dot\z_t(\gamma)=\delta_t\c_t-\M\z_t(\gamma)\qquad\Rightarrow\qquad\z_t=\delta_t\M^{-1}\c_t,
\end{align}
where we have defined $\c_t:=\W\x_t$. Since $\delta_t=0$ when $\x_t$ is a negative sample, the algorithm only produces a non-trivial output when presented with a positive sample. Next, we take a stochastic gradient descent-ascent step with respect to $(\W,\M)$. We can replace the averages $\tfrac1T\Z\X_{(+)}^\top$ and $\tfrac1T\Z\Z^\top$ in equations \eqref{eq:deltaW} and \eqref{eq:deltaM} with the rank-1 approximations $\z_t\x_t^\top$ and $\z_t\z_t^\top$, respectively. To estimate $\B_\beta$ in the online setting, we note that
\begin{align*}
    \C_{(-)}=\langle\x_t\x_t^\top|\delta_t=0\rangle_t=\frac{\langle(1-\delta_t)\x_t\x_t^\top\rangle_t}{\langle1-\delta_t\rangle_t}.
\end{align*}
Therefore, we can replace $\B_\beta$ with the online estimate
\begin{align*}
    \beta\frac{1-\delta_t}{p_t}\x_t\x_t^\top+(1-\beta)\I_d,
\end{align*}
where $p_t$ denotes a running estimate of $\langle1- \delta_t\rangle_t$ that is initialized at $p_0=0.5$ and updated at each time step, as follows:
\begin{align*}
    p_t=p_{t-1}+\frac1t(1-\delta_t-p_{t-1}).
\end{align*}
These substitutions result in the online updates
\begin{align}\label{eq:W}
    \W&\gets\W+2\eta\left(\z_t-\beta\frac{1-\delta_t}{p_t}\c_t\right)\x_t^\top-2\eta(1-\beta)\W\\ \label{eq:M}
    \M&\gets\M+\frac\eta\tau\left(\z_t\z_t^\top-\M\right).
\end{align}
We can map these onto a neural network, Fig.~\ref{fig:nn}.
Assuming that each neuron has access to the running average $p_t$, the learning rules are local in the sense that the update to a synaptic weight depends only on variables that are available in the pre- and postsynaptic neurons.

\section{Numerical experiments}


\subsection{Comparison of cPCA and cPCA\*}

\begin{figure}
    \centering
    \includegraphics{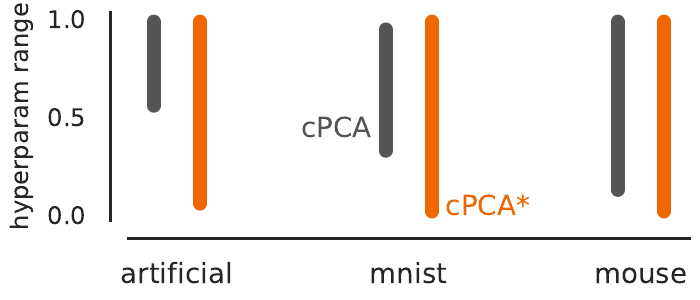}
    \caption{Range of hyper-parameters  ($\alpha$ for cPCA and $\beta$ for cPCA\*) that lead to good performance (greater than 90\% classification accuracy using linear discriminant analysis) is broader in  cPCA\*.}
    \label{fig:barcodes}
\end{figure}

We compare cPCA and cPCA\* on two naturalistic datasets used in \cite{cpca} as  well as a synthetic dataset. The results are summarized in Fig.~\ref{fig:barcodes}.\bigskip

\textbf{Artificial dataset.} The artificial dataset consists of 200 30-dimensional samples. The negative samples are pure Gaussian noise while the positive samples exhibit a bimodal distribution along some of the dimensions. Due to high variance in the noisy directions, PCA is ineffective for identifying the clusters.

\begin{figure}
    \centering\includegraphics{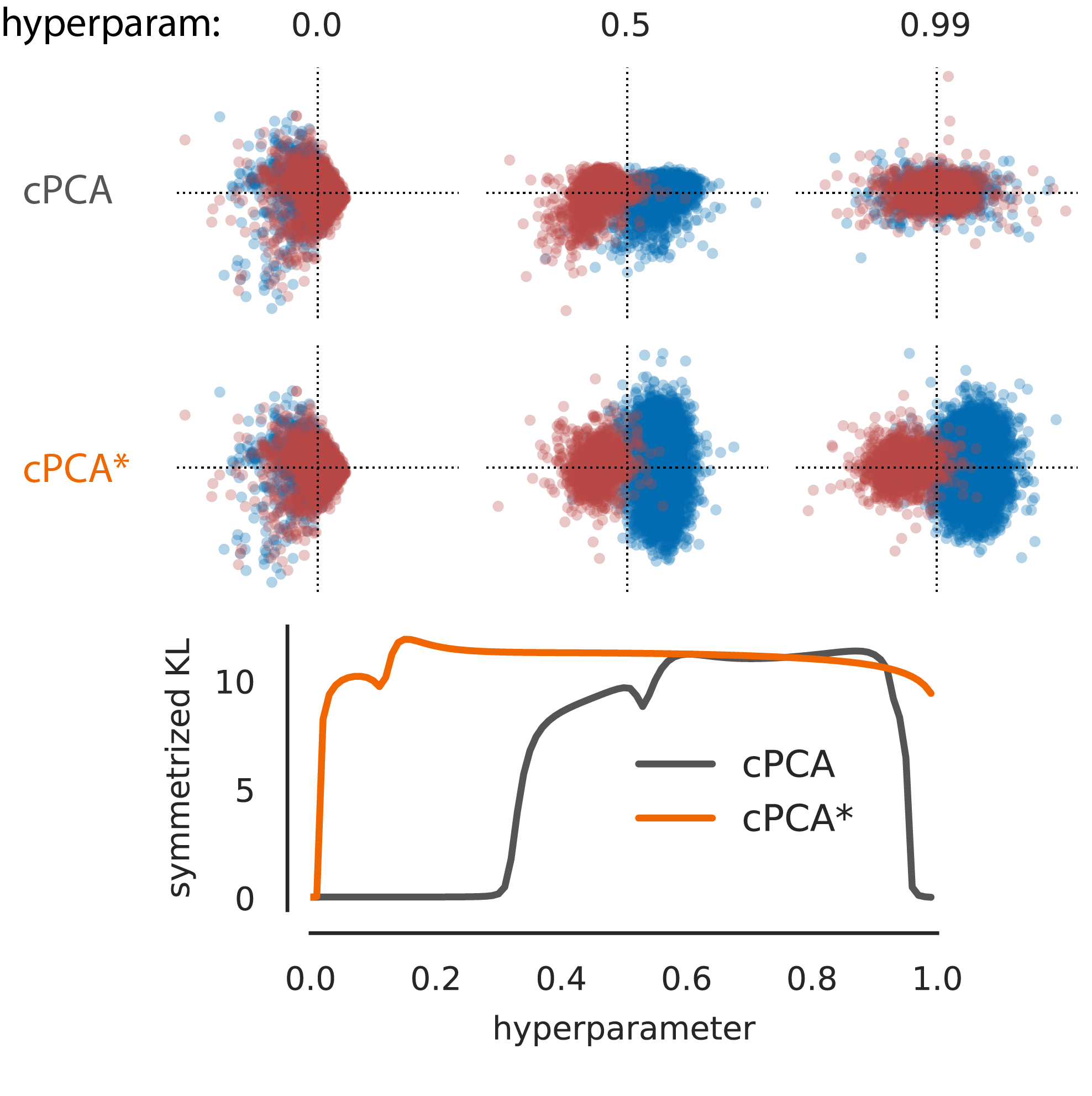}\vspace{-1.5em}
    \caption{cPCA\* is more robust than cPCA to the choice of hyper-parameters. Separation of noisy MNIST digits 0 and 1 with (top) cPCA (top) vs.\ (middle) cPCA\*. Bottom: Quantitative measure of separation of the MNIST digits vs.\ value of hyper-parameter ($\alpha$ for cPCA and $\beta$ for cPCA\*).
    \label{fig:mnist_hyper}}
\end{figure}

\textbf{Noisy Digits dataset.} In the first naturalistic dataset, the positive samples consist of 5,000 synthetic images generated by randomly superimposing images of handwritten digits 0 and 1 from MNIST dataset \cite{lecun1998mnist} on top of 
natural images taken from the UPenn Natural Image Database~\cite{Tkacik2011}. The negative samples consist of natural images. We apply cPCA and cPCA\* to find 2-dimensional projections, for varying hyper-parameter choices, Figs.~\ref{fig:mnist_hyper} and~\ref{fig:barcodes}. To measure how well-separated the projections of the noisy digits 0 and 1 are, we plot the symmetrized KL divergence of the distributions of the projections of the noisy 0's and the distributions of the projections of the noisy 1's. We see that our method separates the digits for a much larger range of the hyper-parameter (Fig.~\ref{fig:mnist_hyper}).

\textbf{Mouse dataset.} The second naturalistic dataset consists of protein expression measurements $(d=77)$ of mice that have received shock therapy~\cite{mouse}, some of which develop Down Syndrome. We apply cPCA and cPCA\* (with $k=2$) using negative samples that consist of protein expression measurements from a set of mice that have not been exposed to shock therapy and measure the degree of linear separability between mice that do not have Down Syndrome and mice that have Down Syndrome, Fig.~\ref{fig:barcodes}. 

\subsection{Evaluation of online cPCA\*}

\begin{figure}
    \centering
    \includegraphics[trim={0 15pt 0 0},width=.4\textwidth]{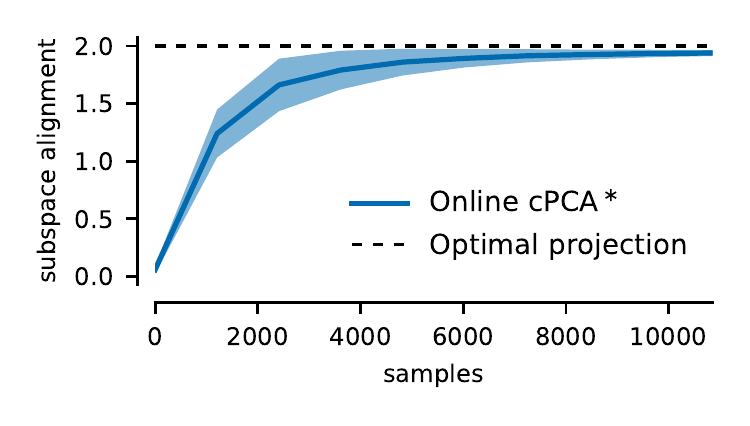}
    \caption{Convergence of the online cPCA\* algorithm on the mouse dataset. The $y$-axis shows the alignment of the projector found via the online algorithm to the optimal projector, computed by taking the trace of the product of the two projectors. Shaded region shows standard deviation over 5 runs of the experiment.}
    \label{fig:online}
\end{figure}

Fig.~\ref{fig:online} shows the convergence of the online cPCA\* algorithm on the mouse dataset. We see that cPCA\* converges to the optimal projector. Here, the learning rate is $\eta=0.003$ and the ratio between gradient descent and ascent steps is $\tau=1$.

\section{Summary}

In this work, we introduced cPCA\*, a modified contrastive Principal Component Analysis method. We showed that this method is interpretable as maximizing the signal to noise ratio and leads to an online algorithm which can be mapped onto a neural network with local learning rules. In terms of the scope of applicability, cPCA\* has the same requirements as cPCA: as a contrastive algorithm it needs a relevant background dataset. However, we hope that the derivation of this online algorithm with more robust hyper-parameter sensitivity will broaden the possible use-cases of this algorithm to larger datasets.


\clearpage

\bibliographystyle{IEEEbib}
\bibliography{biblio}

\end{document}